\documentclass[journal]{IEEEtran}

\usepackage{times}
\usepackage{graphicx} 
\usepackage{subfigure} 

\usepackage{algorithm}
\usepackage{algorithmic}
\usepackage{setspace}

\usepackage{helvet}
\usepackage{courier}
\usepackage{hyperref}

\usepackage{makecell}
\usepackage{graphicx}
\usepackage{subfigure}
\usepackage{color}
\usepackage{amsfonts}
\usepackage{amsmath}
\usepackage{amssymb}

\usepackage{multirow}
\usepackage{booktabs}
\usepackage{cleveref}

\DeclareMathAlphabet\mathbfcal{OMS}{cmsy}{b}{n}

\def\0{{\bf 0}}
\def\1{{\bf 1}}

\usepackage{ntheorem}

\newtheorem*{*thm}{Theorem}

\newtheorem*{*lemma}{Lemma}

\usepackage{multicol}

\usepackage{subfigure}
\usepackage{multirow}
\usepackage{url}
\usepackage{cite}
\usepackage{hyperref}
\usepackage{microtype}
\usepackage{bm}
\usepackage{amssymb}

\usepackage{times}
\usepackage{latexsym}

\usepackage[T1]{fontenc}

\usepackage[utf8]{inputenc}

\usepackage{microtype}

\usepackage{inconsolata}

\usepackage{graphicx}

\usepackage{algorithm}
\usepackage{algorithmic}
\usepackage{color}
\usepackage{listings}

\usepackage{booktabs}

\usepackage{tabularx}
\usepackage{multirow}
\usepackage{amsmath}
\usepackage[table]{xcolor}
\usepackage{amsfonts}
\usepackage{float} 
\usepackage{colortbl}
\usepackage{setspace}
\usepackage[accsupp]{axessibility}  
\title{M2IST: Multi-Modal Interactive Side-Tuning for Efficient Referring Expression Comprehension}

\author{Xuyang Liu, Ting Liu, Siteng Huang, Yi Xin, Yue Hu, Long Qin, Donglin Wang, \textit{Member, IEEE} \\ Yuanyuan Wu, and Honggang Chen, \textit{Member, IEEE}

	\IEEEcompsocitemizethanks{
	\IEEEcompsocthanksitem{Xuyang Liu and Ting Liu contributed equally to this work.
 
    This work was supported in part by the National Natural Science Foundation of China (Grant Nos. 62001316, 62306329, 62103425), the Sichuan Science and Technology Program (Nos. 2024ZYD0263 and 2024YFHZ0212), the Open Foundation of Yunnan Key Laboratory of Software Engineering (2023SE206), and the Fundamental Research Funds for the Central Universities under Grant SCU2023D062, 2022CDSN-15-SCU. \textit{(Corresponding authors: Honggang Chen and Siteng Huang.)}
    
    Xuyang Liu is with the College of Electronics and Information Engineering, Sichuan University, Chengdu 610065, China (email: liuxuyang@stu.scu.edu.cn).

    Ting Liu, Yue Hu, and Long Qin are with with the College of Systems Engineering, National University of Defense Technology, Changsha 410072, China. (email: \{liuting20, huyue11, qinlong\}@nudt.edu.cn).

    Yi Xin is with the State Key Laboratory for Novel Software Technology, Nanjing University, Nanjing 210000, China (email: xinyi@smail.nju.edu.cn).

    Siteng Huang and Donglin Wang is with the School of Engineering, Westlake University, Hangzhou 310030, China (email: siteng.huang@gmail.com, wangdonglin@westlake.edu.cn).

    Yuanyuan Wu is with the College of Computer Science and Cyber Security (Pilot Software College), Chengdu University of Technology, Chengdu 610059, China (email: wuyuanyuan@cdut.edu.cn).
    
    Honggang Chen is with the College of Electronics and Information Engineering, Sichuan University, Chengdu 610065, China, and also with the Yunnan Key Laboratory of Software Engineering, Yunnan University, Kunming 650600, China (e-mail: honggang\_chen@scu.edu.cn).}
    
	}
    }

\begin{document}
\maketitle

\begin{abstract}

Referring expression comprehension (REC) is a vision-language task to locate a target object in an image based on a language expression. Fully fine-tuning general-purpose pre-trained vision-language foundation models for REC yields impressive performance but becomes increasingly costly. Parameter-efficient transfer learning (PETL) methods have shown strong performance with fewer tunable parameters. However, directly applying PETL to REC faces two challenges: (1) insufficient multi-modal interaction between pre-trained vision-language foundation models, and (2) high GPU memory usage due to gradients passing through the heavy vision-language foundation models. To this end, we present M2IST: Multi-Modal Interactive Side-Tuning with M3ISAs: Mixture of Multi-Modal Interactive Side-Adapters. During fine-tuning, we fix the pre-trained uni-modal encoders and update M3ISAs to enable efficient vision-language alignment for REC. Empirical results reveal that M2IST achieves better performance-efficiency trade-off than full fine-tuning and other PETL methods, requiring only 2.11\% tunable parameters, 39.61\% GPU memory, and 63.46\% training time while maintaining competitive performance. Our code is released at \url{https://github.com/xuyang-liu16/M2IST}.

\end{abstract}


\begin{figure*}[t]
\centering
\includegraphics[width=0.95\textwidth]{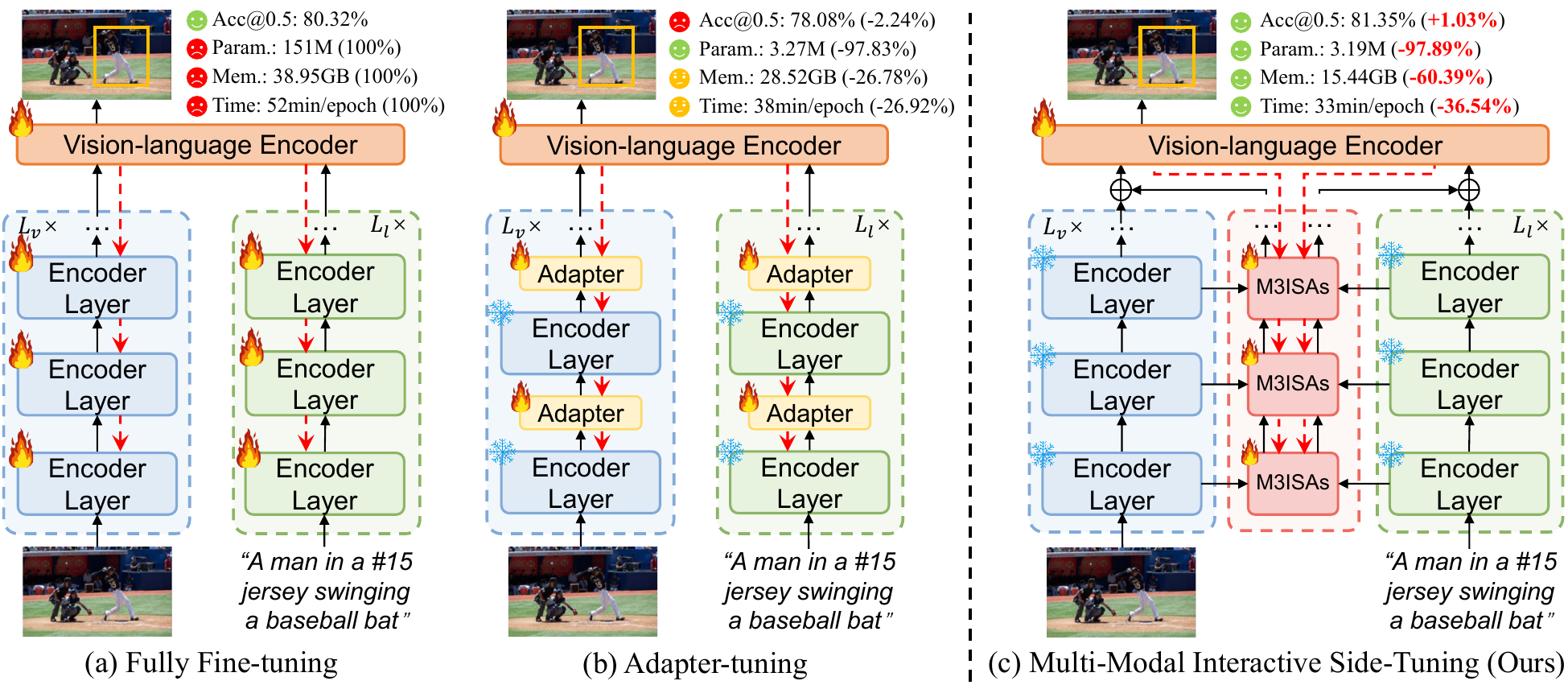}
\vspace{-2mm}
\caption{Comparison of (a) fully fine-tuning, (b) Adapter-tuning, and (c) our M2IST for REC. By only using 2.11\% of the tunable parameters, 39.61\% of the GPU memory, and 63.46\% of the fine-tuning time, M2IST achieves comparable or even superior performance compared to fully fine-tuning.}
\vspace{-4mm}
\label{fig:intro}
\end{figure*}

\IEEEpeerreviewmaketitle

\section{Introduction}
\label{sec:intro}

\IEEEPARstart{R}{eferring} expression comprehension (REC) is one of the most challenging vision-language tasks, aiming to locate a specific object in an image based on a given referring expression \cite{yu2018mattnet,yang2019fast,deng2021transvg,hua2023MRLN,qiu2024mcce,ji2024progressive}. Recent studies \cite{deng2021transvg,sun2022proposal,Cross,shi2023dynamicmdetr} have demonstrated impressive performance by fine-tuning general-purpose vision-language foundation models for this task. Following this trend but with increasingly larger model sizes, early approaches such as TransVG~\cite{deng2021transvg}, Dynamic M-DETR~\cite{shi2023dynamicmdetr} adapt pre-trained DETR~\cite{carion2020detr} and BERT~\cite{devlin2018bert}, while more recent works scale up to larger architectures - MaPPER~\cite{liu-2024-mapper} employs DINOv2~\cite{oquab2024dinov2}, and SimVG~\cite{dai2023simvg} and C$^3$VG~\cite{dai2025C3VG} utilize pre-trained BEiT-3~\cite{beit3} to achieve better performance for REC. Although these approaches demonstrate continuous performance improvements, fully fine-tuning pre-trained models remains computationally expensive when adapting to new datasets (see Fig.~\ref{fig:intro} (a)). Additionally, fine-tuning on limited REC data may lead to catastrophic forgetting and overfitting, as recently evaluated in other vision-language tasks \cite{huang2023vop,yuan2023mrsadapter,jiang2022cmadapter}.

Recently, parameter-efficient transfer learning (PETL) methods \cite{houlsby2019adapter,hu2021lora,jia2022vpt,chen2022adaptformer} have been proposed to address similar issues by updating only a small set of parameters to efficiently adapt pre-trained models to downstream tasks. Adapter-tuning \cite{houlsby2019adapter}, a prominent PETL method, has demonstrated significant success across various vision-language foundation models as well as downstream tasks \cite{yuan2023mrsadapter,xu2023bridging}. It typically inserts a tunable lightweight bottleneck-shaped module sequentially into each frozen backbone layer. Most \textit{transformer-based} REC models \cite{deng2021transvg,sun2022proposal,zhu2022seqtr,wu2024duet,lu2024lgr-net} use pre-trained Vision Encoder and Language Encoder to separately extract image and text features, which are then integrated to form multi-modality features for reasoning. A straightforward approach to apply adapter-tuning for REC is to insert the adapters into the transformer encoder layers to enhance fine-tuning efficiency (see Fig.~\ref{fig:intro} (b)). However, this introduces \textit{\textbf{two significant challenges}}: \textbf{(1)} Updating inserted adapters still requires backpropagation of gradients through the large pre-trained vision and language models, placing a heavy burden on GPU memory during fine-tuning (see Fig.~\ref{fig:intro} (b)). \textbf{(2)} Vision-language foundation models are pre-trained separately, each with its own structure and training data \cite{carion2020detr,devlin2018bert}. Directly inserting vanilla adapters into them may lack cross-modality interaction in the shallow layers of the entire modal, resulting in sub-optimal vision-language alignment. This is particularly problematic for predicting referred objects in cases with complex semantic information, such as human actions and spatial relations (see Fig.~\ref{fig:visualizations} "Without Interaction"). Most recently, a series of memory-efficient transfer learning (METL) methods~\cite{fu2023dtl,mercea2024time,zhang2024quantized,yin2023e3va} have been proposed to reduce GPU memory consumption during fine-tuning. However, these METL methods are primarily designed for single-modality tasks, such as image recognition~\cite{fu2023dtl,mercea2024time}, dense prediction~\cite{yin2023e3va}, and natural language understanding and generation~\cite{zhang2024quantized,hao2024MEFT}. Direct application to REC tasks may still lead to sub-optimal vision-language understanding.

To address these challenges, we propose a novel \textbf{\underline{M}}ulti-\textbf{\underline{M}}odal \textbf{\underline{I}}nteractive \textbf{\underline{S}}ide-\textbf{\underline{T}}uning (\textbf{M2IST}) method that effectively strengthens vision-language alignment and enables parameter- and memory-efficient transfer to REC within the unified \textit{interactive side networks} (see Fig.~\ref{fig:intro} (c)). Specifically, we introduce \textbf{\underline{M}}ixture of \textbf{\underline{M}}ulti-\textbf{\underline{M}}odal \textbf{\underline{I}}nteractive \textbf{\underline{S}}ide-\textbf{\underline{A}}dapters (\textbf{M3ISAs}), which incorporate Vision Expert Adapters (VEA), Language Expert Adapters (LEA), and Interaction Expert Adapters (IEA) into the \textit{side networks} in parallel with the heavy encoders. VEA and LEA transfer pre-trained single-modality knowledge to the REC domain. IEA utilizes a linear layer for weight-sharing between image and text features, enabling progressive 
interaction between the referring sentence and input image. Such interaction aggregates \textbf{\textit{channel-level}} vision-language alignment at shallow layers of the model, facilitating deep \textbf{\textit{token-level}} vision-language alignment in deeper layers for improved performance. This elegant design achieves parameter-, memory-, and time-efficient \textbf{\textit{intra-}} and \textbf{\textit{inter-}}modality representation transfer for REC.

We conduct extensive experiments on RefCOCO \cite{yu2016refcoco}, RefCOCO+ \cite{yu2016refcoco}, and RefCOCOg \cite{mao2016refcocogg,nagaraja2016refcocogu} to demonstrate the effectiveness and efficiency of M2IST for REC. Experimental results show that M2IST achieves the optimal performance-parameter-memory trade-off compared to most full fine-tuning methods and other PETL methods. By applying our M2IST method, a standard transformer-based REC model can requiring only \textbf{2.11\%} of the tunable parameters, \textbf{39.61\%} of the GPU memory and \textbf{63.46\%} of the fine-tuning time compared to full fine-tuning, while still achieving competitive performance (see Fig.~\ref{fig:intro}). With the sufficient vision-language interaction strengthened by our M3ISAs, our method can accurately locate the referred objects for various complex cases (see Fig.~\ref{fig:visualizations}).  

To summarize, our main \textbf{contributions} are three-fold: 
\begin{enumerate}

        \item We propose M2IST, a novel Multi-Modal Interactive Side-Tuning method for referring expression comprehension (REC), which effectively addresses the challenges of insufficient multi-modal interaction and high GPU memory consumption in applying parameter-efficient transfer learning (PETL) to REC.
        
        \item We design Mixture of Multi-Modal Interactive Side Adapters (M3ISAs), seamlessly integrating pre-trained vision and language encoders, enabling parameter-, memory-, and time-efficient tuning within a unified interactive side network for REC.
        
        \item We conduct empirical studies on the application of PETL methods in REC, highlighting their limitations in practical scenarios. Extensive experiments on three widely-used benchmarks validate the effectiveness of M2IST, achieving the optimal trade-off between performance and efficiency compared to full fine-tuning and other PETL methods, with significantly reduced GPU memory usage and fine-tuning time.

\end{enumerate}

The rest of this paper is organized as follows. In Section~\ref{sec:related work}, we briefly review related work on referring expression comprehension and efficient transfer learning. We then provide a detailed description of our proposed M2IST method and its core component, M3ISA, followed by a discussion of the advantages of M2IST in Section~\ref{sec:Methodology}. In Section~\ref{sec:Experiments}, we present extensive quantitative and qualitative experiments to analyze the performance and efficiency of M2IST. Finally, we summarize our work in Section~\ref{sec:Conclusion} and discuss the limitations and future work in Section~\ref{sec:Limiations}.

\section{Related Work}
\label{sec:related work}

\subsection{Referring Expression Comprehension}
Referring expression comprehension (REC) \cite{yu2018mattnet,yang2019fast,deng2021transvg,sun2022proposal,hua2023MRLN,shi2023dynamicmdetr} aims to locate the specific objects in images based on textual descriptions. 

Early REC methods \cite{zhang2018vc,yu2018mattnet,liu2019learning} follow a \textit{two-stage} pipeline that first uses a pre-trained object detector (e.g., Faster R-CNN \cite{ren2015faster}) to generate a set of sparse object proposals, which are then ranked by their similarity to the given textual description. MAttNet \cite{yu2018mattnet} uses Faster R-CNN to generate object proposals, then scores them based on the referring expression to find the most relevant object. RvG-Tree \cite{hong2019learning} builds a relational visual graph to capture object interactions and ranks proposals according to their relevance to the expression, improving grounding accuracy. However, these \textit{two-stage} REC methods heavily rely on the quality of the object proposals and cannot directly predict the referred object region. \textit{One-stage} anchor-based methods \cite{yang2019fast,liao2020real,yang2020improving,ye2021one} have been developed to eliminate the proposal generation step, directly predicting object bounding boxes from predefined anchors \cite{redmon2018yolov3}. FAOA \cite{yang2019fast} utilizes the YOLOv3 detector \cite{redmon2018yolov3}, integrating it with an encoded language vector to ground the referred regions. MRLN \cite{hua2023MRLN} introduces three modules: feature-feature relational learning, feature-task relational learning, and task-task relational learning, to enhance the collaborative learning of REC, effectively reducing prediction inconsistency in multi-task learning. Recently, \textit{transformer-based} methods \cite{deng2021transvg,du2022vgtr,sun2022proposal,shi2023dynamicmdetr} have demonstrated superior performance by implicitly modeling cross-modality relationships within a unified architecture. The pioneering work TransVG \cite{deng2021transvg} applies a stack of transformer encoders to perform feature extraction and multi-modal fusion for REC. VGTR \cite{du2022vgtr} employs a transformer encoder-decoder architecture to jointly reason over visual and textual inputs, grounding the referred object without relying on pre-trained detectors or word embeddings. Dynamic M-DETR \cite{shi2023dynamicmdetr} uses a dynamic multi-modal transformer decoder to adaptively sample visual features and perform text-guided decoding for REC. 

As REC models continue to scale up, their performance has shown significant improvements. However, most \textit{transformer-based} methods, including TransVG~\cite{deng2021transvg}, VGTR~\cite{du2022vgtr}, PFOS~\cite{sun2022proposal}, and Dynamic M-DETR~\cite{shi2023dynamicmdetr}, require \textbf{fully fine-tuning} of their pre-trained networks. While these approaches achieve satisfactory performance, they come with a substantial computational cost, demanding larger GPU memory to accommodate the increased number of trainable parameters (see Fig.~\ref{fig:intro} (a)). To this end, our M2IST aims to provide an \textbf{efficient fine-tuning} strategy for these \textit{transformer-based} REC models, significantly reducing the computational costs during the fine-tuning stage.

\subsection{Parameter-efficient Transfer Learning}
Parameter-efficient transfer learning (PETL) \cite{houlsby2019adapter,hu2021lora,jia2022vpt,chen2022adaptformer} has emerged as a promising alternative to fully fine-tuning pre-trained models for downstream tasks. By updating only a minimal subset of parameters, PETL methods balance performance and computational efficiency. 

Recent PETL methods can be classified into two categories. The first category is updating additional parameters in modules inserted into the model \cite{houlsby2019adapter,chen2022adaptformer} or appended to the input data \cite{jia2022vpt,huang2023vop,shi2024explicit}. Adapter \cite{houlsby2019adapter} incorporates a bottleneck module into each Transformer layer, positioned after both the Multi-Head Attention (MHA) and the Feed-Forward Networks (FFN). AdaptFormer \cite{chen2022adaptformer} embeds the adapter module parallel to the FFN in each encoder of a Vision Transformer. VPT \cite{jia2022vpt} appends learnable visual vectors to the input sequences (VPT-Shallow) or to the input of each transformer encoder layer (VPT-Deep). The second category involves decomposing weight matrices into two low-rank matrices and updating only the small factorization matrices \cite{hu2021lora,jie2023fact}. As a pioneering work, LoRA \cite{hu2021lora} integrates a tunable pair of low-rank decomposed weight matrices into each encoder layer of the pre-trained networks. FacT \cite{jie2023fact} incorporates tunable factorized weight matrices into each layer of the pre-trained networks. There is also growing interest in adapter-based PETL methods for vision-language tasks, including text-image retrieval \cite{yuan2023mrsadapter} and text-video retrieval \cite{jiang2022cmadapter}. Most of these methods aim to achieve effective cross-modality interaction while maintaining parameter efficiency. 

However, existing PETL methods still face substantial GPU memory consumption during the fine-tuning stage, as gradients must propagate through the heavy pre-trained encoders for REC (see Fig.~\ref{fig:intro} (b)). Compared with above PETL methods, our M2IST aims to not only reduce the number of updated parameters but also significantly decrease the required \textbf{GPU memory consumption}, thereby providing a more comprehensive efficient fine-tuning strategy.

\subsection{Memory-efficient Transfer Learning}
Memory-efficient transfer learning (METL) \cite{zhang2020side,sung2022lst,fu2023dtl} aims to reduce memory costs on GPUs during fine-tuning. Existing METL methods typically employ a \textit{side network} for single-modality knowledge transfer, focusing on either NLP \cite{sung2022lst} or CV \cite{fu2023dtl} downstream tasks. Side-Tuning \cite{zhang2020side} utilized an additional side network that adds its representation to the backbone network in the last layer. LST \cite{sung2022lst} adopted a separate and lightweight side network with the same architecture as the pre-trained model but reduced each layer dimension by a pre-defined reduction factor. DTL \cite{fu2023dtl} disentangled the weights update from the pre-trained backbone network by proposing a lightweight side network, which achieved high accuracy in classification with low GPU memory usage. 

While existing MEFT methods, particularly side-tuning approaches, have achieved notable results in reducing GPU memory consumption, they are primarily designed for single-modality tasks and fail to effectively bridge pre-trained vision and language models for the challenging task of REC. In this work, inspired by the existing efforts, our M2IST bridges the pre-trained Vision Encoder and Language Encoder through the unified \textit{side networks}, enabling a parameter-, memory-, and time-efficient transfer to the REC task (see Fig.~\ref{fig:intro} (c)).

\begin{figure*}[t]
\centering
\includegraphics[width=\textwidth]{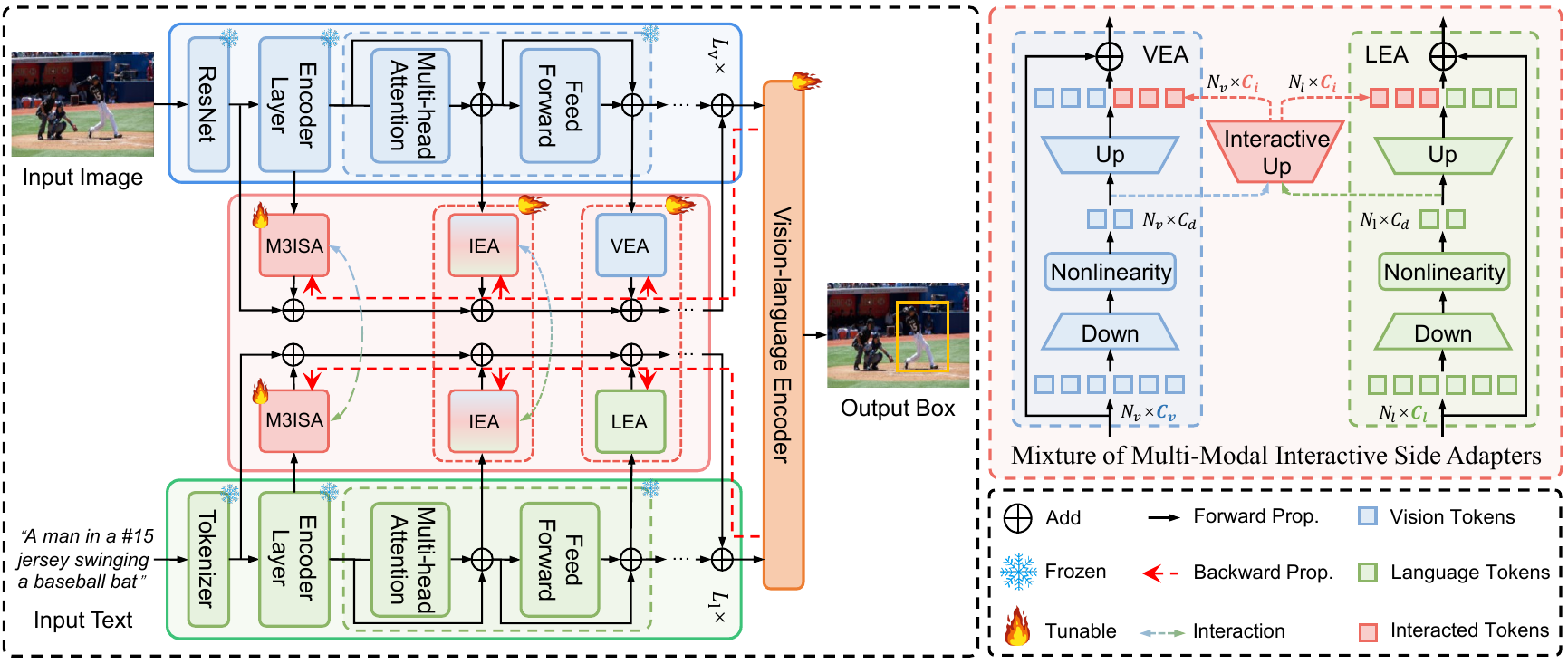}
\vspace{-4mm}
\caption{\textbf{Overall architecture of M2IST.} M2IST freezes the pre-trained Vision Encoder (\textbf{\textcolor[rgb]{0.180, 0.459, 0.714}{blue branch}}) and Language Encoder (\textbf{\textcolor[rgb]{0.439, 0.678, 0.278}{green branch}}), while updating M3ISAs on \textit{side networks} (\textbf{\textcolor[rgb]{0.929, 0.420, 0.380}{pink branch}}). M3ISAs comprise IEA for bridging the pre-trained dual encoders to enable cross-modality interactions, and VEA/LEA for transferring pre-trained single-modality representations to adapt to the REC domain. By avoiding backpropagation through the heavy encoders (\textbf{\textcolor{red}{red dashed arrow}}), M2IST enables \textbf{parameter-, memory-, and time-efficient} tuning for the task of referring expression comprehension.}
\vspace{-4mm}
\label{fig:overview}
\end{figure*}

\section{Methodology}
\label{sec:Methodology}

In this section, we present our M2IST in detail. First, we briefly overview the base architecture for referring expression comprehension in Section~\ref{sec:Base}. Then, we elaborate on the designs of our efficient tuning method M2IST and its core component M3ISA in Section~\ref{sec:M2IST}. Finally, we provide an in-depth analysis of some advantages of M2IST in Section~\ref{sec:Advantages}.

\subsection{Base Architecture}
\label{sec:Base}

We apply a standard \textit{transformer-based} REC model as our base architecture, shown in Fig.~\ref{fig:intro} (a), which comprises: (1) a Vision Encoder, (2) a Language Encoder, and (3) a Vision-language Encoder.

\paragraph{Vision Encoder} We adopt a DETR-based \cite{carion2020detr} encoder as our Vision Encoder, which comprises a ResNet \cite{he2016resnet} and a stack of transformer encoder layers to encode the image into high-quality vision embeddings. Specifically, given an input image $\bm{z}_0 \in \mathbb{R}^{H_0 \times W_0 \times 3}$, the ResNet is utilized to generate a 2D feature map $\bm{z} \in \mathbb{R}^{H \times W \times C}$, where $H_0$ and $W_0$ denote the height and width of the input image, $H=\frac{H_0}{32}$, $W=\frac{W_0}{32}$, and $C=2048$ represents the channel dimension. Then, a $1\times 1$ convolutional layer reduces the $C$ to $C_v=256$, producing $\bm{z}' \in \mathbb{R}^{H \times W \times C_v}$. We flatten the feature map $\bm{z}'$ into a sequence of 1D vectors (i.e., vision tokens) $\bm{z}_v \in \mathbb{R}^{N_v \times C_v}$, where $N_v=H\times W$ indicates the number of tokens. Sequentially, these vision tokens added with positional encodings are fed into a stack of 6 transformer encoder layers, which then output the enhanced vision embeddings $\bm{f}_v \in \mathbb{R}^{N_v \times C_v}$ incorporating global context of the image.

\paragraph{Language Encoder} We employ an off-the-shelf language model BERT \cite{devlin2018bert}, comprising a stack of transformer encoder layers, as our Language Encoder. Specifically, given the input text, each word ID is converted into a one-hot vector, which is then tokenized into a sequence of language tokens. These language tokens, concatenated with a \texttt{[CLS]} token at the beginning and a \texttt{[SEP]} token at the end, are input to 12 transformer encoder layers to sequentially model contextual relationships. Similar to the Vision Encoder, Language Encoder finally outputs the enhanced language embeddings $\bm{f}_l \in \mathbb{R}^{N_l \times C_l}$, where $N_l$ and $C_l=768$ represent the number and channel dimension of language tokens, respectively.

\paragraph{Vision-language Encoder} We use a transformer-based encoder \cite{vaswani2017attention} as our Vision-language Encoder (V-L Encoder) to thoroughly fuse the multi-modality embeddings and predict the bounding box of the referred object. Specifically, the enhanced vision embeddings $\bm{f}_v \in \mathbb{R}^{N_v \times C_v}$ and language embeddings $\bm{f}_l \in \mathbb{R}^{N_l \times C_l}$ are first projected into the joint embeddings $\bm{f'}_v \in \mathbb{R}^{N_v \times C_p}$ and $\bm{f'}_l \in \mathbb{R}^{N_l \times C_p}$, sharing the same channel dimension $C_p=256$. The joint embeddings, along with a learnable \texttt{[REG]} token, are then fed into a stack of 6 transformer encoder layers to fuse the cross-modality embeddings and output the \texttt{[REG]} token. Finally, a prediction head, implemented as a Multi-layer Perceptron with two 256-dim hidden layers and a linear output layer, receives the \texttt{[REG]} token and regresses it to the 4-dim box coordinates for the referred object.

\subsection{Multi-Modal Interactive Side-Tuning: M2IST}
\label{sec:M2IST}

Given that the pre-trained Vision Encoder and Language Encoder contain rich knowledge and comprise about 95\% of the model's parameters. We first explore two approaches to reduce training overhead:
\begin{enumerate}
    \item \textbf{Fully freezing the pre-trained encoders.} We choose to directly keep the pre-trained parameters fixed and only fine-tune the V-L Encoder. While it effectively saves a significant amount of GPU memory, it also results in significantly inferior performance (see Table~\ref{Table:ablation on components} (a)). 
    \item \textbf{Updating a few additional parameters.} We explore various mainstream PETL methods, such as Adapter \cite{houlsby2019adapter} and LoRA \cite{hu2021lora}. Though most of them achieve relatively satisfactory performance as well as save tunable parameters, updating the additional parameters still necessitates substantial GPU memory rather than effectively mitigating the computational load (see Table~\ref{Table:comparisons with PETL}). 
\end{enumerate} 

To address the above issues, we propose Multi-Modal Interactive Side-Tuning (M2IST) that keeps the pre-trained encoders frozen and updates the proposed Mixture of Multi-Modal Interactive Side Adapters (M3ISA) on \textit{side networks} to facilitate parameter- and memory-efficient fine-tuning for REC, as shown in Fig.~\ref{fig:overview}. Note that we do not show the LayerNorm for simplicity.

\paragraph{M3ISA Architecture} The core component of our M2IST is M3ISA (see Fig.~\ref{fig:overview} (right)), which consists of two distinct adapters, i.e., \textbf{intra-} and \textbf{inter-}modality adapters to effectively and efficiently bridge the pre-trained Vision Encoder and Language Encoder.

The intra-modality adapters include \textbf{\textit{Vision Expert Adapter}} (VEA) and \textbf{\textit{Language Expert Adapter}} (LEA), shown as separate \textbf{\textcolor[rgb]{0.180, 0.459, 0.714}{blue branch}} and \textbf{\textcolor[rgb]{0.439, 0.678, 0.278}{green branch}} in Fig.~\ref{fig:overview} (right). They adhere to a fundamental design \cite{houlsby2019adapter}  for transferring pre-trained single-modality representations to more domain-specific ones. Specifically, both of them consist of a linear down-projection layer $\mathbf{W}_{\text{down}}$, ReLU non-linear activation, and a linear up-projection layer $\mathbf{W}_{\text{up}}$ in sequence. Taking the VEA as a formulaic example, given the vision tokens $\bm{x}_v \in \mathbb{R}^{N_v \times C_v}$, the function of VEA can be formally expressed as:
\begin{equation}
\text{VEA}(x_v) = x_v + s\cdot{\text{ReLU}(x_v\mathbf{W}_{\text{down}})}\mathbf{W}_{\text{up}},
\end{equation}
where $\mathbf{W}_{\text{down}} \in \mathbb{R}^{C_v \times C_d}$, $\mathbf{W}_{\text{up}} \in \mathbb{R}^{C_d \times C_v}$ and $s$ is the scaling factor of the adapter.

In the base architecture, the Vision Encoder and Language Encoder are responsible for extracting single-modality features, while only the V-L encoder is tasked with cross-modality fusion through \textbf{\textit{token-level}} vision-language alignment. However, such the late fusion has been proved insufficient when the referring sentence contains complex semantic information, such as spatial relationships \cite{huang2023JMCELN}. Therefore, we introduce an inter-modality adapter, named the Interaction Expert Adapter (IEA). Specifically, each IEA shares a set of tokens within the same channel dimensions $C_i$ (i.e., Interacted Tokens in Fig.~\ref{fig:overview} (right)) between the Vision Encoder and Language Encoder, thereby achieving \textbf{\textit{channel-level}} vision-language alignment during the multi-modal feature extraction stage. As depicted by the \textbf{\textcolor[rgb]{0.929, 0.420, 0.380}{entire pink section}} in Fig.~\ref{fig:overview} (right), IEA include a unique down-projection layer for vision \textcolor[rgb]{0.180, 0.459, 0.714}{$\mathbf{W}_{\text{down}}$} $\in \mathbb{R}^{\textcolor[rgb]{0.180, 0.459, 0.714}{C_v} \times C_d}$ and language \textcolor[rgb]{0.439, 0.678, 0.278}{$\mathbf{W}_{\text{down}}$} $\in \mathbb{R}^{\textcolor[rgb]{0.439, 0.678, 0.278}{C_l} \times C_d}$, ReLU activation, an interactive up-projection layer \textcolor[rgb]{0.929, 0.420, 0.380}{$\mathbf{W}_{\text{up}}$} $\in \mathbb{R}^{C_d \times \textcolor[rgb]{0.929, 0.420, 0.380}{C_i}}$, and a unique up-projection layer for vision \textcolor[rgb]{0.180, 0.459, 0.714}{$\mathbf{W}_{\text{up}}$} $\in \mathbb{R}^{C_d \times (\textcolor[rgb]{0.180, 0.459, 0.714}{C_v}-\textcolor[rgb]{0.929, 0.420, 0.380}{C_i})}$ and language \textcolor[rgb]{0.439, 0.678, 0.278}{$\mathbf{W}_{\text{up}}$} $\in \mathbb{R}^{C_d \times (\textcolor[rgb]{0.439, 0.678, 0.278}{C_l}-\textcolor[rgb]{0.929, 0.420, 0.380}{C_i})}$, where \textcolor[rgb]{0.180, 0.459, 0.714}{$C_v$}, \textcolor[rgb]{0.439, 0.678, 0.278}{$C_l$}, and \textcolor[rgb]{0.929, 0.420, 0.380}{$C_i$} represent the vision, language, and interaction channels, respectively. Given the vision tokens $\bm{x}_v \in \mathbb{R}^{N_v \times \textcolor[rgb]{0.180, 0.459, 0.714}{C_v}}$ and language tokens $\bm{x}_l \in \mathbb{R}^{N_l \times \textcolor[rgb]{0.439, 0.678, 0.278}{C_l}}$, the corresponding down-projection layers first down-sample them to the bottleneck features $\bm{z}_v \in \mathbb{R}^{N_v \times C_d}$ and $\bm{z}_l \in \mathbb{R}^{N_l \times C_d}$.
Then, the corresponding up-projection layers and interactive up-projection layer up-sample these bottleneck features and concatenate them within the same modality to obtain the cross-modality features $\bm{f}_v \in \mathbb{R}^{N_v \times \textcolor[rgb]{0.180, 0.459, 0.714}{C_v}}$ and $\bm{f}_l \in \mathbb{R}^{N_l \times \textcolor[rgb]{0.439, 0.678, 0.278}{C_l}}$ as:
\begin{equation}
f_v = \text{Concat}[z_v\textcolor[rgb]{0.180, 0.459, 0.714}{\mathbf{W}_{\text{up}}}, z_v\textcolor[rgb]{0.929, 0.420, 0.380}{\mathbf{W}_{\text{up}}}],
\end{equation}
\vspace{-0.4cm}
\begin{equation}
f_l = \text{Concat}[z_l\textcolor[rgb]{0.439, 0.678, 0.278}{\mathbf{W}_{\text{up}}}, z_l\textcolor[rgb]{0.929, 0.420, 0.380}{\mathbf{W}_{\text{up}}}].
\end{equation}
The outputs of the IEA can be written as:
\begin{equation}
\text{IEA}(x_v) = x_v + s\cdot{f_v}, 
\end{equation}
\vspace{-0.4cm}
\begin{equation}
\text{IEA}(x_l) = x_l + s\cdot{f_l},
\end{equation}
where $x_v$ and $x_l$ indicate input as vision tokens and language tokens, respectively. 

\begin{algorithm}[t]
\tiny
\caption{PyTorch-like pseudocode for IEA and M2IST.}
\definecolor{codeblue}{rgb}{0.25,0.5,0.5}
\definecolor{colorred}{RGB}{197, 49, 124}
\label{alg:code}
\lstset{
  backgroundcolor=\color{white},
  basicstyle=\fontsize{8.8pt}{8.8pt}\ttfamily\selectfont,
  columns=fullflexible,
  breaklines=true,
  captionpos=b,
  commentstyle=\fontsize{7.2pt}{7.2pt}\color{codeblue},
  keywordstyle=\fontsize{7.2pt}{7.2pt}\color{colorred},
}
\begin{lstlisting}[language=python]
# Define the IEA Module
class IEA(nn.Module):
    def __init__(self, vis_dim_model, text_dim_model, down_bottle, share_up, inter_up):
        self.text_down_proj = nn.Linear(text_dim_model, down_bottle)
        self.vis_down_proj = nn.Linear(vis_dim_model, down_bottle)
        self.inter_up = nn.Linear(down_bottle, share_up)
        self.text_up_proj = nn.Linear(down_bottle, text_dim_model)
        self.vis_up_proj = nn.Linear(down_bottle, vis_dim_model)

    def forward(self, text_x, vis_x):
        vis_down = self.vis_down_proj(vis_x)
        text_down = self.text_down_proj(text_x)
        vis_inter_up = self.inter_up(vis_down)
        text_inter_up = self.inter_up(text_down)
        vis_up =  self.vis_up_proj(vis_down)
        text_up = self.text_up_proj(text_down)
       
        return concat[text_up,text_inter_up], concat[vis_up,vis_inter_up]

IEA_out, LEA_out, VEA_out = [], [], []

# Multi-Modal Interactive Side-Tuning
for i in range(layers):
    IEA_out = IEA(text_mha_output,vis_mha_output)
    LEA_out = LEA(text_ffn_out)
    VEA_out = VEA(vis_ffn_out)
    
    IEA_out.append(IEA_out)
    LEA_out.append(LEA_out)
    VEA_out.append(VEA_out)

final_text_feature = text_feature + sum(IEA_out) + sum(LEA_out)
final_vis_feature = vis_feature + sum(IEA_out) + sum(VEA_out)

\end{lstlisting}
\end{algorithm}

Our IEA enables lightweight adaptation of cross-modality representations without increasing the explicit computational burden of vision-language feature extraction, thereby efficiently bridging the pre-trained vision and language encoders. Our experimental results in Table~\ref{Table:ablation on hyper-parameters} further demonstrate that deeper \textbf{\textit{channel-level}} vision-language alignment yields better performance for REC. Moreover, such alignment will further improve the \textbf{\textit{token-level}} vision-language alignment in V-L Encoder, leading to better region-sentence understanding in challenging REC scenarios, as analyzed in Section~\ref{sec:Qualitative}.

\paragraph{M3ISA Implementation} As depicted in Fig.~\ref{fig:overview} (left), we incorporate a stack of M3ISAs into two \textit{side networks} that operate \textit{in parallel with} the pre-trained dual encoders. Specifically, in one encoder layer (both for vision and language), the IEA first receives processed vision/language tokens from the Multi-head Attention (MHA) layers as input and produces adapted, interacted tokens for the vision/language side network. Subsequently, the VEA/LEA take the processed vision/language tokens from the Feed Forward Networks (FFN) as input and generate adapted single-modality tokens for the corresponding side networks. The outputs of the IEA and VEA/LEA are added within the vision/language side networks, along with the original vision/language tokens through skip-connections. After passing through the side networks, the outputs of the vision/language side networks are added to the outputs of the vision/language encoders. During fine-tuning, we keep the pre-trained encoders fixed and update the M3ISAs in the \textit{side networks}, allowing the pre-trained encoders to act as \textit{standalone feature extractors}. We present the PyTorch-like pseudocode of the proposed IEA module and M2IST in Algorithm~\ref{alg:code} for better understanding.

\paragraph{Training Objectiveness} Following most \textit{transformer-based} REC methods \cite{deng2021transvg,shi2023dynamicmdetr}, the training loss function is a combination of the widely used smooth L1 loss and GIoU loss. Specifically, the prediction is donated as $\mathbf{b}=(x, y, w, h)$, and the normalized ground-truth box as $\hat{\mathbf{b}}=(\hat{x},\hat{y},\hat{w},\hat{h})$. The training objective is:
\begin{equation}
    \mathcal{L} = \mathcal{L}_{\text{smooth-l1}}(\mathbf{b}, \hat{\mathbf{b}}) + \lambda \cdot \mathcal{L}_{\text{giou}}(\mathbf{b}, \hat{\mathbf{b}}),
\end{equation}
where $\mathcal{L}_{\text{smooth-l1}}(\cdot)$ and $\mathcal{L}_{\text{giou}}(\cdot)$ are the smooth L1 loss and GIoU loss. $\lambda$ is the weight coefficient of GIoU loss to balance these two losses.

\begin{table*}[!t]

\caption{\textbf{Comparison with efficient tuning methods using the same base architecture.} "Param." indicates the number of tunable parameters in the pre-trained encoders. "Mem." denotes the peak GPU memory footprint with \textbf{batch size 64} during fine-tuning. \\ We highlight the \textbf{best} and the \underline{second} results.}
\centering

\small
\setlength{\tabcolsep}{6.5pt}

\begin{tabular}{l|cc|ccc|ccc|ccc}
\toprule
    
\multirow{2}{*}{Methods} & \multicolumn{1}{c}{Params.$\downarrow$} & \multicolumn{1}{c|}{Mem.$\downarrow$} & \multicolumn{3}{c|}{RefCOCO} & \multicolumn{3}{c|}{RefCOCO+} & \multicolumn{3}{c}{RefCOCOg} \\

 & (M) & (GB) & val & testA & testB & val & testA & testB & val-g & val-u & test-u \\ \midrule

\textcolor{gray}{Fully fine-tuning} & \textcolor{gray}{151 (100\%)} & \textcolor{gray}{38.95 (100\%)} & \textcolor{gray}{80.32} & \textcolor{gray}{82.67} & \textcolor{gray}{75.24} & \textcolor{gray}{63.50} & \textcolor{gray}{68.15} & \textcolor{gray}{55.63} & \textcolor{gray}{66.56} & \textcolor{gray}{67.66} & \textcolor{gray}{67.44} \\ \midrule


\textbf{\textit{PETL Methods:}} & & & & & & & & & & \\

Adapter \cite{houlsby2019adapter} & 3.27 (2.17\%) & 28.52 (73.22\%) & 78.02 & 79.89 & 75.23 & 61.35 & 66.34 & 54.21 & 63.18 & 65.26 & \underline{66.65} \\

LoRA \cite{hu2021lora} & \underline{2.37 (1.57\%)} & 20.37 (52.30\%) & 77.57 & 78.22 & 73.37 & 61.24 & \underline{66.53} & 53.95 & 64.27 & \underline{67.36} & 66.43 \\

AdaptFormer \cite{chen2022adaptformer} & 2.38 (1.57\%) & 20.37 (52.30\%) & 76.32 & 77.16 & 73.94 & 60.96 & 65.19 & 53.88 & 61.81 & 65.44 & 64.37 \\

CM Adapter \cite{jiang2022cmadapter} & 3.27 (2.17\%)& 27.19 (69.81\%) & 77.37 & 78.81  & 74.07 & 61.34 & 66.10 & 53.31 & 63.93 & 65.75 & 64.72 \\ 

MRS-Adapter \cite{yuan2023mrsadapter} & \textbf{1.58 (1.05\%)} & 20.07 (51.53\%) & 77.14 & 77.80 & 74.80 & 61.13 & 66.38 & 53.13 & 63.07 & 66.46 & 65.16 \\


\midrule

\textbf{\textit{METL Methods:}} & & & & & & & & & & \\

DTL \cite{fu2023dtl} & 3.22 (2.13\%) & \underline{15.65 (40.18\%)} & \underline{79.89} & \underline{80.52} & \underline{76.33} & \underline{62.14} & 66.41 & \underline{54.49} & \underline{65.31} & 66.23 & 65.94 \\

\rowcolor{gray!20}




\textbf{M2IST (Ours)} & 3.19 (2.11\%) & \textbf{15.44 (39.64\%)} & \textbf{81.35} & \textbf{82.29}  & \textbf{77.98}  & \textbf{63.15} & \textbf{67.11} & \textbf{55.52} & \textbf{67.50} & \textbf{67.67} & \textbf{67.41} \\

\bottomrule

\end{tabular}

\vspace{-2mm}
\label{Table:comparisons with PETL}
\end{table*}

\subsection{Discussion: Advantages of M2IST}
\label{sec:Advantages}

The proposed M2IST offers several advantages over fully fine-tuning and other PETL methods for REC, which we can summarize as \textbf{\textit{three efficiency factors}}:

\begin{enumerate}

  \item \textbf{Parameter Efficiency.} Fully fine-tuning pre-trained vision-language foundation models is computationally expensive due to their large size and complexity \cite{qiu2024mcce,liu2024vgdiffzero}. Furthermore, it often leads to forgetting valuable pre-trained knowledge and increases the risk of overfitting, as the encoders are fine-tuned on limited data. M2IST mitigates these issues by freezing the pre-trained encoders and updating only the lightweight M3ISAs, achieving effective intra- and inter-modality representation adaptation and enhanced performance.
  
  \item \textbf{Memory Efficiency.} Both full fine-tuning and other PETL methods require backpropagation through large pre-trained vision-language foundation models, leading to high GPU memory usage \cite{sung2022lst,fu2023dtl}. M2IST reduces this by separating tunable parameters from the pre-trained encoders and placing them in parallel \textit{side interactive networks}. Since gradients backpropagate through the lightweight M3ISAs instead of the heavy encoders, where we only need to store the parameters of our side networks, this leads to reduced GPU memory requirements. Additionally, M2IST maintains the baseline model's architecture, simplifying its implementation compared to other PETL methods.
    
  \item \textbf{Time Efficiency.} Most PETL methods are able to enhance the speed of fine-tuning compared to full fine-tuning, primarily due to the reduced number of updated parameters. Compared to other PETL methods, our M2IST may offer greater both fine-tuning and inference time efficiency. As for the fine-tuning stage, M2IST introduces tunable parameters in parallel with the pre-trained encoders, allowing gradient computation to occur in the lightweight M3ISAs instead of the computationally intensive encoders. As for the inference stage, compared to most PETL methods that introduce new parameters into the pre-trained networks, compromising inference efficiency, our M2IST can process the additional parameters and the pre-trained networks in parallel, making the inference more efficient.

\end{enumerate}

Therefore, our approach offers parameter-, memory-, and time-efficient adaptation for visual-language foundational models, and enhances the REC task with more comprehensive visual-language alignment.

\section{Experiments}
\label{sec:Experiments}

In this section, we first introduce the datasets and evaluation metrics used to assess the performance and efficiency of our method in Section~\ref{sec:Setup}. Next, we describe the implementation details in Section~\ref{sec:more imple}. Section~\ref{sec:Comparison} provides comprehensive comparisons of performance and fine-tuning efficiency, followed by an extensive ablative study and analysis of design choices in Section~\ref{sec:Ablation}. Finally, we present multiple qualitative results to analyze model behavior in Section~\ref{sec:Qualitative}.

\subsection{Datasets and Evaluation Metrics}
\label{sec:Setup}

\paragraph{Datasets} To verify the effectiveness and efficiency of our method, we conduct experiments on the following REC benchmarks as follows: (1) \textit{RefCOCO} \cite{yu2016refcoco} consists of 19,994 images with 142,210 referring expressions for 50,000 objects. The RefCOCO dataset is officially split into train, validation, testA, and testB sets containing 120,624, 10,834, 5,657, and 5,095 expressions, respectively. (2) \textit{RefCOCO+} \cite{yu2016refcoco} includes 19,922 images with 141,564 referring expressions for 49,856 objects. Compared to RefCOCO, the referring expressions in RefCOCO+ focus more on attributes of the referred objects, such as color and shape, without including any positional words. (3) \textit{RefCOCOg} \cite{mao2016refcocogg,nagaraja2016refcocogu} contains 25,799 images with 95,010 referring expressions for 49,822 objects. Compared to RefCOCO and RefCOCO+, the referring expressions in RefCOCOg are typically longer, averaging almost twice the length of those in the other two datasets. RefCOCOg has two commonly used split strategies: the \textit{google} split \cite{mao2016refcocogg} (-g) and the \textit{umd} split \cite{nagaraja2016refcocogu} (-u). 

\paragraph{Evaluation Metrics} Following previous work \cite{deng2021transvg,shi2023dynamicmdetr}, 
we conduct experiments on both RefCOCOg-g (val-g) and RefCOCOg-u (val-u and test-u). We use Precision@0.5 as the evaluation metric. In addition to accuracy, we also report the number of tunable parameters in the pre-trained encoders and the peak training memory consumption in Gigabytes (GB) to compare the fine-tuning efficiency with other PETL methods. Specifically, we consider the number of parameters in full fine-tuning as the baseline (100\%) and evaluate the parameter efficiency of other methods as the percentage of their tunable parameters relative to full fine-tuning. Similarly, we consider the memory utilization of full fine-tuning as the baseline (100\%) and quantify the memory efficiency of other efficient tuning methods as the ratio between their peak memory usage and that of full fine-tuning during the training phase.

\subsection{Implementation Details}
\label{sec:more imple}

\paragraph{Model Weights} The Vision Encoder is initialized with the backbone (i.e., ResNet-50 \cite{he2016resnet}) and encoder weights from DETR \cite{carion2020detr}, which is pre-trained on the entire MS-COCO dataset \cite{lin2014microsoftcoco}. Specifically, during the pre-training of the Vision Encoder, images from the validation and test sets of RefCOCO/+/g that overlap with MS-COCO \cite{lin2014microsoftcoco} are excluded. The Language Encoder is initialized with BERT-base \cite{devlin2018bert}, pre-trained on the BookCorpus \cite{zhu2015BookCorpus} and English Wikipedia \cite{devlin2018bert}. The Vision-Language (V-L) Encoder is initialized using Xavier initialization. The proposed M3ISAs are initialized with Kaiming normal initialization.

\paragraph{Hyper-parameters Settings} M3ISAs are inserted into the transformer encoder layers at the same indices as those in the Vision Encoder and Language Encoder, and relevant ablation study is conducted in Table~\ref{Table:ablation on position}. Unless otherwise specified, the bottleneck dimensions of the Visual Expert Adapter (VEA) and Language Expert Adapter (LEA) $C_d$ are set to 128 by default, while the interaction dimension $C_i$ of the Interaction Expert Adapter (IEA) is 256 by default. The scaling factor $s$ for all adapters is set to 0.1.

\paragraph{Training Details} For RefCOCO \cite{yu2016refcoco} and RefCOCOg \cite{mao2016refcocogg,nagaraja2016refcocogu} datasets, the entire network is trained for 90 epochs using the AdamW optimizer, with a learning rate of $10^{-4}$ for the V-L Encoder and $10^{-5}$ for the M3ISAs. The weight decay is $10^{-4}$, and the learning rate is reduced by a factor of 10 after 60 epochs. While for RefCOCO+ \cite{yu2016refcoco}, the network is trained for 180 epochs with the same learning rates and weight decay, but the learning rate is decreased by a factor of 10 after 120 epochs. All experiments are conducted on one A800 GPU.

\subsection{Main Results}
\label{sec:Comparison}

In this sub-section, we compare our M2IST under different settings to evaluate its performance and the \textbf{\textit{three efficiency factors}} discussed in Section~\ref{sec:Advantages}.

\begin{table*}[!t]

\caption{\textbf{Comparison with full fine-tuning on RefCOCO, RefCOCO+, and RefCOCOg.} "RN50" and "RN101" represent ResNet-50 \cite{he2016resnet}, ResNet-101 \cite{he2016resnet}. "Params." is the number and average percentage of tuned parameters in the encoders.}
\centering

\small
\setlength{\tabcolsep}{4.7pt}

\begin{tabular}{l|cc|c|ccc|ccc|ccc}
\toprule
    
\multirow{2}{*}{Methods} & \multicolumn{1}{c}{Vision} &\multicolumn{1}{c|}{Language} & \multicolumn{1}{c|}{Params.$\downarrow$} & \multicolumn{3}{c|}{RefCOCO} & \multicolumn{3}{c|}{RefCOCO+} & \multicolumn{3}{c}{RefCOCOg} \\

 & Encoder & Encoder & (M) & val & testA & testB & val & testA & testB & val-g & val-u & test-u \\ \midrule


\textbf{\textit{Two-stage:}} & & & & & & & & & & & & \\
VC \cite{zhang2018vc} & VGG16 & LSTM & 17 (100\%) & - & 73.33 & 67.44 & - & 58.40 & 53.18 & 62.30 & - & - \\
ParalAttn \cite{zhuang2018paralattn} & VGG16 & LSTM & 17 (100\%) & - & 75.31 & 65.52 & - & 61.34 & 50.86 & 58.03 & - & - \\
MAttNet \cite{yu2018mattnet} & RN101 & LSTM & 47 (100\%) & 76.65 & 81.14 &  69.99 & 65.33 & \textbf{71.62} & 56.00 & - & 66.58 & 67.27 \\
RvG-Tree \cite{hong2019learning} & RN101 & LSTM & 47 (100\%) & 75.06 & 78.61 & 69.85 & 63.51 & 67.45 & 56.66 & - & 66.95 & 66.51 \\
\midrule

\textbf{\textit{One-stage:}} & & & & & & & & & & & & \\
FAOA \cite{yang2019fast} & DarkNet-53 & LSTM & 43 (100\%) & 72.54 & 74.35 & 68.50 & 56.81 & 60.23 & 49.60 & 56.12 & 61.33 & 60.26 \\
RCCF \cite{liao2020rccf} & DLA34 & LSTM & 18 (100\%) & - & 81.06 & 71.85 & - & 70.35 & 56.32 & - & - & 65.73 \\
ReSC \cite{yang2020improving} & DarkNet-53 & BERT & 152 (100\%) & 76.59 & 78.22 & 73.25 & 63.23 & 66.64 & 55.53 & 63.12 & 67.30 & 67.20 \\ 
RealGIN \cite{zhou2021realgin} & DarkNet-53 & GRU & 41 (100\%) & 77.25 & 78.70 & 72.10 & 62.78 & 67.17 & 54.21 & - & 62.75 & 62.33 \\
TRAR \cite{zhou2021trar} & DarkNet-53 & LSTM & 43 (100\%) & - & 79.60 & 71.30 & - & 65.10 & 53.50 & - & 63.30 & 62.50 \\
TransVG \cite{deng2021transvg} & RN50+DETR & BERT & 151 (100\%) & 80.32 & 82.67 & 75.24 & 63.50 & 68.15 & 55.63 & 66.56 & 67.66 & 67.44 \\
VGTR \cite{du2022vgtr} & RN50 & LSTM & 52 (100\%) & 78.70 & 82.09 & 73.31 & 63.57 & 69.65 & 55.33 & 62.88 & 65.62 & 65.30 \\
PFOS \cite{sun2022proposal} & DarkNet-53 & BERT & 152 (100\%) & 77.37 & 80.43 & 72.87 & 63.74 & 68.54 & 55.84 & 61.46 & 67.08 & 66.35 \\
DMRNet \cite{zhang2023dmrnet} & DarkNet-53 & BERT & 152 (100\%) & 76.99 & 79.71 & 72.67 & 61.58 & 66.60 & 54.00 & - & 66.03 & 66.70 \\
MRLN \cite{hua2023MRLN} & VGG16 & GRU & 18 (100\%) & \underline{81.39} & \textbf{83.65} & 75.03 & \textbf{66.33} & 69.75 & \textbf{58.05} & - & 65.52 & 65.08 \\
D-MDETR \cite{shi2023dynamicmdetr} & RN50+DETR & BERT & 143 (100\%) & 80.47 & 82.63 & 75.96 & \underline{65.52} & \underline{70.41} & \underline{57.68} & \underline{66.87} & \textbf{69.20} & \textbf{68.56} \\

\midrule
\textbf{\textit{Ours:}} & & & & & & & & & & & & \\
\rowcolor{gray!20}
\textbf{M2IST ($C_d=8$)} & RN50+DETR & BERT & \textbf{1.91 (1.26\%)} & \textbf{81.55}  & \underline{83.07}  & 77.31  & 62.73 & 66.96 & 55.93 & 65.47 & 66.79 & 66.30 \\
\rowcolor{gray!20}
\textbf{M2IST ($C_d=32$)} & RN50+DETR & BERT & \underline{ 2.20 (1.46\%)} & 81.03  & 82.34  & \underline{77.54}  & 62.48  & 66.73 & 55.70 & 66.22 & 67.42  & \underline{67.83} \\
\rowcolor{gray!20}
\textbf{M2IST ($C_d=128$)} & RN50+DETR & BERT & 3.19 (2.11\%) & 81.35 & 82.29  & \textbf{77.98}  & 63.15 & 67.11 & 55.52 & \textbf{67.50} & \underline{67.67} & 67.41 \\
\bottomrule

\end{tabular}
\vspace{-2mm}

\label{Table:comparisons with SOTA}
\end{table*}

\paragraph{Comparison with Efficient Tuning Methods} We first compare our proposed M2IST method with several existing parameter-efficient transfer learning (PETL) methods, including vanilla Adapter \cite{houlsby2019adapter}, LoRA \cite{hu2021lora}, AdaptFormer \cite{chen2022adaptformer}, CM Adapter \cite{jiang2022cmadapter}, and MRS-Adapter \cite{yuan2023mrsadapter}, as well as one memory-efficient transfer learning (METL) method, DTL \cite{fu2023dtl}. All experiments use the same base architecture and maintain consistent bottleneck dimensions ($C_d = 128$).

From Table~\ref{Table:comparisons with PETL}, we can see that all efficient tuning methods including M2IST achieve significant parameter efficiency during fine-tuning by reducing the number of tunable parameters compared to full fine-tuning. Despite the parameter efficiency achieved by these PETL methods, they still face two major limitations in practical applications: a) All methods exhibit performance degradation compared to full fine-tuning. b) They require substantial GPU memory during fine-tuning, thus failing to deliver the efficiency that is crucial during implementation. 

As for the performance, M2IST stands out as the only efficient tuning method that can achieve performance on par with or even better than full fine-tuning, significantly outperforming other methods across all three benchmarks. This highlights the effectiveness of M3ISAs in adapting pre-trained knowledge for the REC task. We can observe that all PETL and METL methods exhibit distinct performance degradation compared to full fine-tuning on the RefCOCOg dataset. This is because RefCOCOg is a substantial dataset with sufficient data, reducing the likelihood of overfitting when fully fine-tuning the models. Even so, with the facilitation of cross-modality interaction between the encoders, M3ISAs are able to enhance the modeling of complex spatial relationships, leading to competitive performance with full fine-tuning on the RefCOCOg benchmark. 

Regarding fine-tuning efficiency, METL methods DTL and our M2IST demonstrate superior memory efficiency compared to other PETL methods. This results from the fact that gradients backpropagate through the lightweight networks rather than the computationally intensive encoders, significantly reducing GPU memory consumption, thus highlighting M2IST's advantage in \textbf{memory efficiency}, as mentioned in Section~\ref{sec:Advantages}.
Besides, our M2IST outperforms DTL due to the fact that its core component M3ISA is able to effectively bridge pre-trained vision-language knowledge, enabling better handling of complex vision-language understanding scenarios for REC.

\begin{table}[t]
\caption{\textbf{Comparisons of training and inference speed.} "Full" represents fully fine-tuning the baseline modal. "Training" denotes training on RefCOCO, while "Inference" denotes inference time on RefCOCO val.}
\centering

\small
\setlength{\tabcolsep}{2.5pt}

\begin{tabular}{l|ccc}
\toprule
Speed & Full  & Adapter & \cellcolor{gray!20}\textbf{M2IST} \\
\midrule
Training (min/epoch) $\downarrow$ & 52 (100\%) & 38 (70.08\%) & \cellcolor{gray!20}\textbf{33 (63.46\%)} \\
Inference (s) $\downarrow$ & \textbf{145 (100\%)} & 146 (100.69\%) & \cellcolor{gray!20}\textbf{145 (100\%)} \\
\bottomrule
\end{tabular}

\vspace{-2mm}
\label{Table:time}
\end{table}

\paragraph{Comparison with Full Fine-tuning Methods} We further compare our M2IST with traditional full fine-tuned REC methods in Table~\ref{Table:comparisons with SOTA}.

Table~\ref{Table:comparisons with SOTA} demonstrates that M2IST achieves competitive performance across three benchmarks compared to most full fine-tuning methods with the fewest (only 1.91\%) trainable backbone parameters. Specifically, on the three sets of RefCOCO \cite{yu2016refcoco}, M2IST outperforms the majority of other fully fine-tuning REC methods. We can observe that MAttNet \cite{yu2018mattnet} and MRLN \cite{hua2023MRLN} achieve impressive performance in in RefCOCO+ \cite{yu2016refcoco}. As a \textit{Two-stage} REC method, MAttNet introduces modular attention networks to separately model the subject, location, and relationship, which can more explicitly locate the referred objects by directly computing the similarity scores between the region proposals and the sentences, thus leading to enhanced performance in RefCOCO+. Similarly, the multiple relational learning modules in MRLN are also well-suited to capture these structured and compositional representations of the vision-language interactions. This allows MRLN to excel on the RefCOCO and RefCOCO+ , where the referring expressions tend to have a more well-defined structure and can be more effectively represented through the learned feature-feature, feature-task, and task-task relationships. However, as the referring expressions become longer and more complex (i.e., RefCOCOg \cite{mao2016refcocogg,nagaraja2016refcocogu}), the limitations of methods that rely heavily on structured representation learning (MAttNet and MRLN) become more apparent. In contrast, transformer-based approaches (e.g., TransVG \cite{deng2021transvg}, D-MDETR \cite{shi2023dynamicmdetr}) that learn end-to-end vision-language representations show promising performance in these more challenging scenarios. Our proposed M2IST further strengthens the vision-language representation learning by combining the channel-level and token-level vision-language alignment, thus achieving impressive performance on the RefCOCOg benchmark.

In summary, Table~\ref{Table:comparisons with SOTA} illustrates that M2IST achieves an optimal performance-efficiency trade-off compared to listed full fine-tuning methods, underscoring its advantage in \textbf{parameter efficiency}, as discussed in Section~\ref{sec:Advantages}.

\paragraph{Comparison of Time Efficiency} We present the training and inference speeds on RefCOCO dataset of full fine-tuning, standard adapter-tuning \cite{houlsby2019adapter}, and our M2IST in Table \ref{Table:time}, where all methods are evaluated on a single A800 GPU. For training speed, following common practice in \textit{transformer-based} methods~\cite{deng2021transvg,shi2023dynamicmdetr}, we train for 90 epochs on RefCOCO with a batch size of 64, and measure the average training time per epoch (min/epoch). For inference speed, we compare the inference time (s) on the RefCOCO validation set.

For training speed, it is evident that PETL methods improve training speed compared to full fine-tuning. This improvement is largely due to the reduced number of parameters that need to be updated when using PETL methods. Notably, M2IST achieves greater efficiency than adapter-tuning in terms of training time by updating parameters in parallel with pre-trained encoders, achieving approximately 36.54\% faster than full fine-tuning. For inference speed, while adapter-tuning slightly reduces inference speed, our M2IST maintains it. This is attributed to the fact that, during inference, M2IST operates in parallel with the pre-trained encoders, maintaining inference speed. In contrast, standard adapters are inserted within the pre-trained encoders, leading to lower computational efficiency. These findings are consistent with the advantage in \textbf{time efficiency} proposed in Section~\ref{sec:Advantages}.

\begin{table}[t]
\caption{\textbf{Comparison on ReferItGame.} Both the standard Adapter and our M2IST use the same base architecture.}
\centering

\small
\setlength{\tabcolsep}{7pt}

\begin{tabular}{l|cc|cc}
\toprule

\multirow{2}{*}{Methods} & \multicolumn{1}{c}{Parameters$\downarrow$} & \multicolumn{1}{c|}{Memory $\downarrow$} & \multicolumn{2}{c}{ReferItGame} \\
& (M) & (GB) & val & test \\ \midrule
Adapter & 3.27 (2.17\%)	& 28.52 (73.22\%) & 57.28 & 56.89 \\
\rowcolor{gray!20}
\textbf{M2IST} & \textbf{3.19 (2.11\%)} & \textbf{15.44 (39.64\%)} & \textbf{60.61} & \textbf{59.30} \\

\bottomrule

\end{tabular}

\vspace{-2mm}
\label{Table:referitgame}
\end{table}

\begin{figure*}[!t]
\centering
\includegraphics[width=0.95\textwidth]{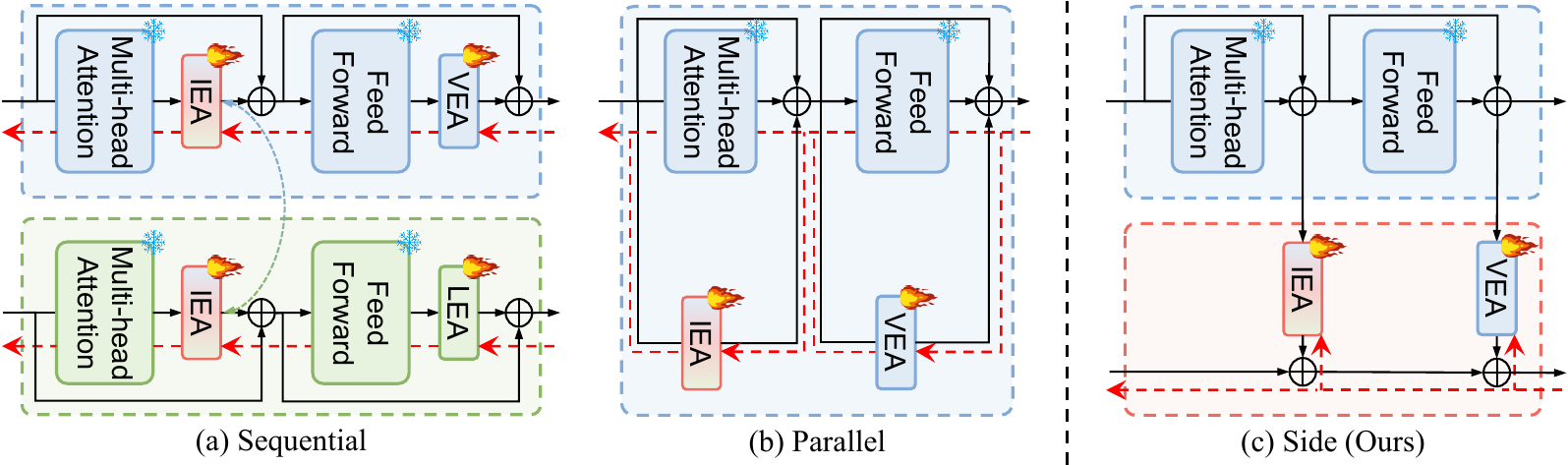}
\vspace{-2mm}
\caption{\textbf{Different adapter insertion forms.} During fine-tuning, gradients in (a) and (b) backpropagate through the heavy encoders, while gradients in (c) only backpropagate through the lightweight adapters, achieving memory-efficient tuning for REC. Note that (b) and (c) only illustrate the vision branch for simplicity.}
\vspace{-4mm}
\label{fig:adapter insertion}
\end{figure*}

In summary, M2IST is Pareto-optimal in terms of accuracy, parameter efficiency, memory efficiency, and time efficiency. By tuning only 3.19M encoder parameters (\textbf{2.11\%} of fully fine-tuning) and requiring 15.44GB of GPU memory (\textbf{39.61\%} of fully fine-tuning) and \textbf{63.46\%} of the full fine-tuning time, M2IST makes fine-tuning a strong REC model on a single NVIDIA 3060 GPU (16GB).

\paragraph{Comparison on Phrase Grounding Task} To demonstrate the generalizability of our M2IST, we have conducted experiments with M2IST on the task of phrase grounding. 

Phrase grounding is a challenging vision-language task that given a noun phrase, output the corresponding single or multiple object detection bounding boxes. If an entire sentence is input, then it's about localizing all the noun phrases included in the sentence. Here, we use the same base architecture in our paper for phrase grounding on ReferItGame dataset \cite{kazemzadeh-etal-2014-referitgame}. Table \ref{Table:referitgame} demonstrates the generalization ability of our M2IST in understanding multi-object scenarios, while also exhibiting significant parameter and memory efficiency during fine-tuning.

\begin{table}[!t]
\caption{\textbf{Ablation on different components in M$^3$ISA}. Without adding any component of M$^3$ISA, it can be viewed as freezing the pre-trained encoder and only training the V-L Encoder.}
\centering

\small
\setlength{\tabcolsep}{3.8pt}

\begin{tabular}{l|ccc|cc|ccc}
\toprule
    
 \multirow{2}{*}{\#} & \multirow{2}{*}{LEA} & \multirow{2}{*}{VEA} & \multirow{2}{*}{IEA} & \multicolumn{1}{c}{Params.$\downarrow$} & \multicolumn{1}{c|}{Mem.$\downarrow$} & \multicolumn{3}{c}{RefCOCO} \\

 &  &  &  & (M) & (GB) & val & testA & testB \\ \midrule

(a) &  &  &  & 0 & 14.32 & 72.72 & 73.33 & 71.27 \\ 
\midrule
(b) & \checkmark &  &  & 0.59 & 14.90 & 77.08 & 77.82 & 73.38 \\
(c) &  & \checkmark &  & 1.02 & 14.52 & 78.30 & 78.95 & 73.58 \\
(d) &  \checkmark & \checkmark &  & 1.61 & 15.09 & 79.39 & 79.18 & 74.41 \\
(e) &  &  & \checkmark & 1.58 & 14.84 & 78.85 & 79.01 & 73.87 \\
\rowcolor{gray!20}
(f) & \checkmark & \checkmark & \checkmark & 3.19 & 15.44 & \textbf{81.35}  & \textbf{82.29}  & \textbf{77.98}
\\
\bottomrule

\end{tabular}

\vspace{-2mm}
\label{Table:ablation on components}
\end{table}

\subsection{Ablation Study and Analysis}
\label{sec:Ablation}

In this sub-section, we investigate the impact of various factors in M2IST. All experiments in this section are performed on three sets of RefCOCO \cite{yu2016refcoco} dataset.

\paragraph{Effects of Different Components of M3ISA} We present the efficiency and performance of various components of M3ISA to examine their effects, as shown in Table~\ref{Table:ablation on components}.

We can see that: \textbf{(1)} Freezing the encoders and only training the V-L Encoder leads to much greater performance degradation (Table~\ref{Table:ablation on components} (a)), indicating a significant domain gap between the pre-trained domains of the two encoders and the REC domain. \textbf{(2)} Fine-tuning single-modality adapters (LEA/VEA) significantly enhances performance compared to using frozen encoders (Table~\ref{Table:ablation on components} (b,c)). Specifically, VEA provides greater performance improvement compared to LEA, suggesting that adapting visual representation plays a more crucial role in object perception and localization than language representation. \textbf{(3)} Combining LEA and VEA yields similar performance to using IEA alone (Table~\ref{Table:ablation on components} (d,e)). This indicates that using either can bring around 6\% accuracy improvement compared to freezing the encoders. \textbf{(4)} Incorporating LEA, VEA, and IEA into M3ISA results in an average improvement of 8.10\% across the three sets of RefCOCO, achieving the best performance among these ablation variants (Table~\ref{Table:ablation on components} (f)). It is worth noting that fine-tuning each ablation variant of M3ISA incurs at most an additional \textbf{1.12GB} of GPU memory compared to freezing the encoder, demonstrating the \textbf{memory efficiency} of M2IST (see Section~\ref{sec:Advantages}).

\begin{table}[!t]
\caption{\textbf{Effects of different mixing strategies of M$^3$ISA}. "VEA+LEA" and "IEA+IEA" refer to adopting the intra-modality adapters and the inter-modality adapters, respectively.}
\centering

\small
\setlength{\tabcolsep}{2.4pt}

\begin{tabular}{l|cc|cc|ccc}
\toprule
    
 \multirow{2}{*}{\#} & \multicolumn{1}{c}{Multi-head} & \multicolumn{1}{c|}{Multi-layer} &  \multicolumn{1}{c}{Params.$\downarrow$} & \multicolumn{1}{c|}{Mem.$\downarrow$} & \multicolumn{3}{c}{RefCOCO} \\

 & Attention & Perceptron & (M) & (GB) & val & testA & testB \\ \midrule

\multicolumn{8}{c}{\textit{Same adapters mixing}} \\ 
(a) & LEA+VEA & LEA+VEA & 3.22 & 15.65 & 79.87 & 80.52 & 76.33 \\ 
(b) & IEA+IEA & IEA+IEA & 3.17 & 14.84 & 78.72 & 80.05 & 76.01
 \\ \midrule
 \multicolumn{8}{c}{\textit{Different adapters mixing}} \\ 
(c) & LEA+VEA & IEA+IEA & 3.19 & 15.38 & 80.58 & 81.26 & 76.65
 \\ 
 \rowcolor{gray!20}
(d) & IEA+IEA & LEA+VEA & 3.19 & 15.44 & \textbf{81.35}  & \textbf{82.29}  & \textbf{77.98}
\\
\bottomrule
\end{tabular}

\vspace{-2mm}
\label{Table:ablation on mixing}
\end{table}

\paragraph{Effects of Different Mixing Strategies of M3ISA} In Table~\ref{Table:ablation on mixing}, to further investigate the effects of different adapter combination forms (i.e., mixing strategies), we present the performance of adopting intra-modality adapters or inter-modality adapters in parallel with different pre-trained layers (MHA and FFN). 

The findings are as follows: \textbf{(1)} Transferring pre-trained single-modality knowledge to the REC domain (e.g., LEA+VEA) is more effective in accurately locating the referred object than merely achieving cross-modality interaction (e.g., IEA+IEA) (Table~\ref{Table:ablation on mixing} (a,b)). \textbf{(2)} Combining intra-modality adapters and inter-modality adapters enhances performance, indicating that joint transfer of pre-trained single-modality knowledge and cross-modality interaction aids in accurately localizing referred objects by text descriptions (Table~\ref{Table:ablation on mixing} (a,b,c,d)). This observation aligns with findings from other challenging vision-language tasks \cite{xu2023bridging,Cross}, suggesting that combining deep inter-modality fusion with intra-modality adaptation improves performance. \textbf{(3)} The best performance among the M3ISA variants is achieved by first connecting the vision and language encoders with IEAs, and then adapting the interacted features and single-modality features to the REC domain with VEA and LEA (Table~\ref{Table:ablation on mixing} (a,b,c,d)).

\paragraph{Effects of Different Insertion Forms of M3ISA} As depicted in Fig.~\ref{fig:adapter insertion} and Table~\ref{Table:ablation on insertion}, we evaluate the impact of integrating M3ISAs with different insertion forms on performance and GPU memory usage. 

From Table~\ref{Table:ablation on insertion}, we can observe that: \textbf{(1)} Side insertion yields the best performance. We suppose that implementing M3ISAs on \textit{side networks} enhances the alignment between the referring sentence and the referred object, resulting in improved localization performance. \textbf{(2)} In terms of fine-tuning efficiency, all three insertion forms contribute to a reduction in GPU memory usage to varying degrees. It is evident that incorporating M3ISAs into the \textit{side networks} consumes the least amount of GPU memory. This is because the gradients backpropagate through the lightweight M3ISAs instead of heavy encoders. This aligns with the \textbf{memory efficiency} advantage mentioned in Section~\ref{sec:Advantages}.

\begin{table}[!t]
\caption{\textbf{Effects of different insertion forms of M3ISA}. "Sequential" and "Parallel", and "Side" correspond to (a), (b), and (c) in Figure~\ref{fig:adapter insertion}, respectively.}
\centering

\small
\setlength{\tabcolsep}{6pt}

\begin{tabular}{l|c|cc|ccc}
\toprule

\multirow{2}{*}{\#} & \multicolumn{1}{c|}{Insertion} & \multicolumn{1}{c}{Params.$\downarrow$} & \multicolumn{1}{c|}{Mem.$\downarrow$} & \multicolumn{3}{c}{RefCOCO} \\

 & forms & (M) & (GB) & val & testA & testB \\ \midrule
(a) & Sequential & 3.19 & 27.19 & 78.76 & 80.25 & 74.90 
\\
(b) & Parallel & 3.19 & 20.37 &  78.29 & 78.71 & 75.30 
\\
\rowcolor{gray!20}
(c) & Side & 3.19 & \textbf{15.44} & \textbf{81.35}  & \textbf{82.29}  & \textbf{77.98} 
\\
\bottomrule

\end{tabular}

\vspace{-2mm}
\label{Table:ablation on insertion}
\end{table}
\begin{table}[t]
\caption{\textbf{Effects of Different Insertion Positions of M3ISA}. The Vision Encoder and Language Encoder consist of 6 and 12 transformer encoder layers, respectively. "$1 \rightarrow 6$" denotes the addition of M3ISAs in the 1st through 6th  encoder layers.}
\centering

\small
\setlength{\tabcolsep}{8.4pt}

\begin{tabular}{l|cc|ccc}
\toprule

\multirow{2}{*}{\#} & \multicolumn{1}{c}{Vision} & \multicolumn{1}{c|}{Language} & \multicolumn{3}{c}{RefCOCO} \\

 & Encoder & Encoder & val & testA & testB \\ \midrule
(a) & $1 \rightarrow 6$ & $1 \rightarrow 6$ & 80.65 & 81.86 & 77.39 
\\
(b) & $1 \rightarrow 6$ & \{1,3,5,7,9,11\}  & 80.83  & 81.76  &  77.54
\\
\rowcolor{gray!20}
(c) & $1 \rightarrow 6$ & $7 \rightarrow 12$  & \textbf{81.35}  & \textbf{82.29}  & \textbf{77.98} 
\\
\bottomrule

\end{tabular}

\vspace{-2mm}
\label{Table:ablation on position}
\end{table}

\paragraph{Effects of Different Insertion Positions of M3ISA} As illustrated in Table~\ref{Table:ablation on position}, we further investigate the impact of introducing M3ISAs at different positions within the pre-trained Vision Encoder and Language Encoder. 

The Vision Encoder and Language Encoder consist of 6 and 12 transformer encoder layers, respectively, and the IEA needs to be inserted into the encoder layers at the same indices. We explore three common insertion forms, as shown in Table~\ref{Table:ablation on position} (a-c). It is evident that inserting our M3ISAs in parallel to the deeper encoder layers of the pre-trained Language Encoder results in better performance. We suggest that deeper encoder layers contain richer semantic features, and establishing cross-modality interaction on this basis helps the model learn finer region-text alignment, thereby achieving better localization performance. Therefore, we adopt insertion form (c) in practical application to achieve the optimal performance.

\begin{table}[t]
\caption{\textbf{Effects of Different Insertion Density of M3ISA.} "Num." indicates M3ISA insertion count, where "0" means only updating V-L Encoder, "2" and "4" denote insertions at layers \{1,4\} and \{1,3,4,6\}, and "6" is the default setting.}
\centering

\small
\setlength{\tabcolsep}{7.0pt}

\begin{tabular}{l|c|cc|ccc}
\toprule

\multirow{2}{*}{\#} & \multirow{2}{*}{Num.} & \multicolumn{1}{c}{Params.$\downarrow$} & \multicolumn{1}{c|}{Mem.$\downarrow$} & \multicolumn{3}{c}{RefCOCO} \\
& & (M) & (GB) & val & testA & testB \\ \midrule
(a) & 0 & 0 & 14.32 & 72.72 & 73.33 & 71.27 \\
\midrule
(b) & 2 & 1.06 & 14.50 & 80.24 & 81.41 & 75.96 \\
(c) & 4 & 2.13 & 14.92 & 80.77 & 81.72 & 77.32 \\
\rowcolor{gray!20}
(d) & 6 & 3.19 & 15.44 & \textbf{81.35} & \textbf{82.29} & \textbf{77.98} \\

\bottomrule

\end{tabular}

\vspace{-2mm}
\label{Table:insert_nums}
\end{table}

\begin{figure*}[t]
\centering
\includegraphics[width=\textwidth]{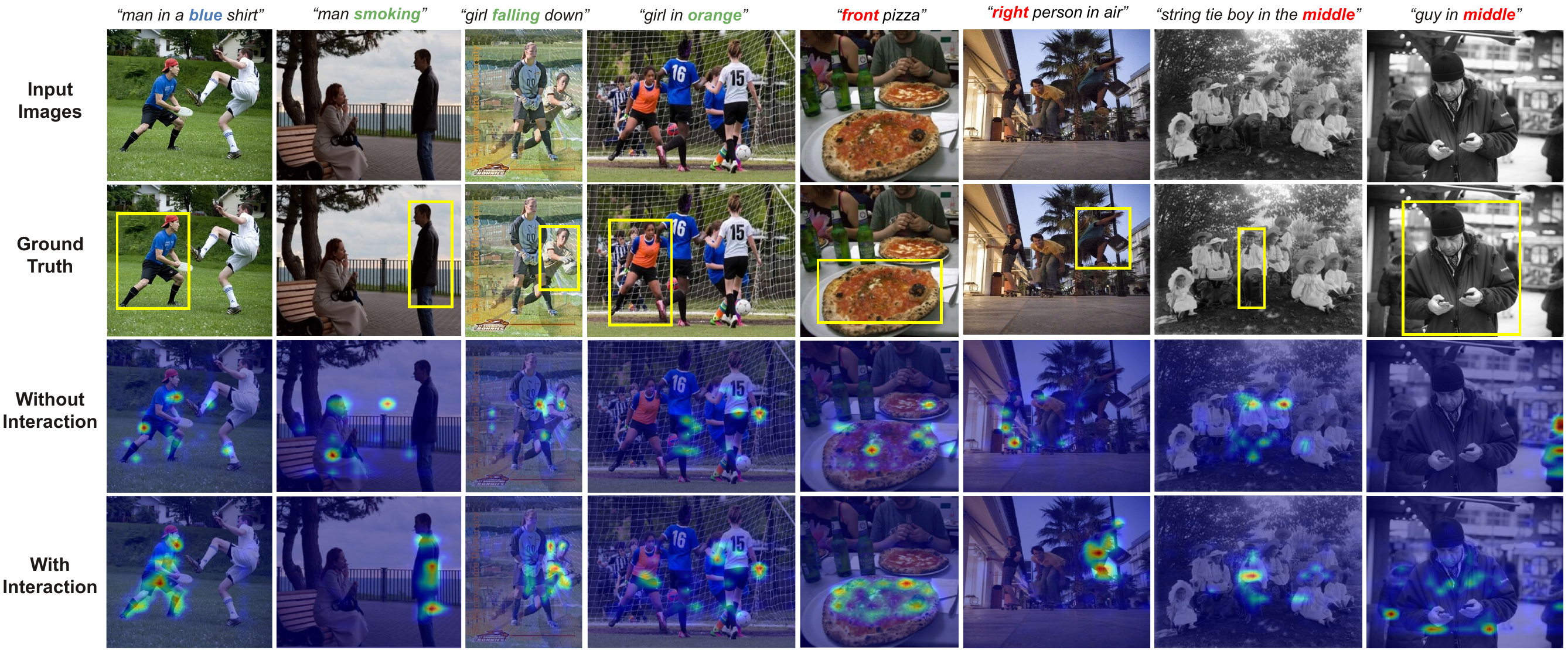}
\vspace{-2mm}
\caption{Visualizations of attention maps from the V-L Encoder with different mixing strategies. Cases include object appearance attributes (\textbf{\textcolor[rgb]{0.180, 0.459, 0.714}{blue words}}), human actions (\textbf{\textcolor[rgb]{0.439, 0.678, 0.278}{green words}}), and spatial relations (\textbf{\textcolor{red}{red words}}).}
\vspace{-2mm}
\label{fig:visualizations}
\end{figure*}

\paragraph{Effects of Different Insertion Density of M3ISA} We further explore the effects of M3ISA insertion density by placing M3ISAs in selected transformer encoder layers.

As presented in Table~\ref{Table:insert_nums}, from (a) and (b), we observe that incorporating M3ISA significantly improves model fine-tuning performance demonstrating effectiveness of M3ISA in adapting pre-trained vision and language models for REC tasks. Moreover, from (b,c,d), It is evident that denser integration of M3ISA modules into pre-trained networks consistently improves model performance without significantly increasing GPU memory consumption during fine-tuning, demonstrating the \textbf{memory efficiency} of our M2IST (see Section~\ref{sec:Advantages}). We adopt the default configuration of six M3ISA insertions to achieve optimal performance.

\begin{table}[t]
\caption{\textbf{Effects of different interaction dimensions of M3ISA.} "$C_i$" represents the interaction dimension of Interaction Expert Adapter (IEA) in M3ISAs.}
\centering

\small
\setlength{\tabcolsep}{7.7pt}

\begin{tabular}{l|c|cc|ccc}
\toprule

\multirow{2}{*}{\#} & \multirow{2}{*}{$C_i$} & \multicolumn{1}{c}{Params.$\downarrow$} & \multicolumn{1}{c|}{Mem.$\downarrow$} & \multicolumn{3}{c}{RefCOCO} \\
&  & (M) & (GB) & val & testA & testB \\ \midrule

(a) & 64 & 2.00 & 15.34 & 77.31 & 77.87 & 73.27 \\
(b) & 128 & 2.40 & 15.35 & 79.26 & 79.58 & 74.60 \\
\rowcolor{gray!20}
(c) & 256 & 3.19 & 15.44 & \textbf{81.35}  & \textbf{82.29}  & \textbf{77.98}  \\

\bottomrule

\end{tabular}

\vspace{-2mm}
\label{Table:ablation on hyper-parameters}
\end{table}









\begin{table}[t]
\caption{\textbf{Effects of combining M2IST with LoRA.} "Full" represents fully fine-tuning the baseline modal. "M2IST+LoRA" represents combining M2IST with LoRA.}
\centering

\small
\setlength{\tabcolsep}{6.6pt}

\begin{tabular}{l|cc|ccc}
\toprule

\multirow{2}{*}{Methods} & \multicolumn{1}{c}{Params.$\downarrow$} & \multicolumn{1}{c|}{Mem.$\downarrow$} & \multicolumn{3}{c}{RefCOCO} \\
 & (M) & (GB) & val & testA & testB \\ \midrule
Full & 151 & 38.95 & 80.32 & 82.67 & 75.24 \\
LoRA & 2.37 & 20.37 & 77.57 & 78.22 & 73.37 \\
\rowcolor{gray!20}
M2IST & \textbf{3.19} & \textbf{15.44} & 81.35 & 82.29 & \textbf{77.98} \\
M2IST+LoRA & 6.46 & 21.68 & \textbf{81.83} & \textbf{82.83} & 77.54 \\

\bottomrule

\end{tabular}

\vspace{-2mm}
\label{Table:mist+lora}
\end{table}

\paragraph{Effects of Different Interaction Dimensions of M3ISA}
We further ablate the impact of changing the interaction dimensions $C_i$ of inter-modality adapters (i.e., IEA), and follow the paradigm of Table~\ref{Table:ablation on mixing} (d).

As depicted in Table~\ref{Table:ablation on hyper-parameters}, deeper \textit{channel-level} vision-language alignment in IEA provides better cross-modality interaction, thus resulting in an increase in performance. When $C_i$ is set to 256 to achieve the optimal trade-off among accuracy, number of tunable parameters, and GPU memory consumption. It is worth noting that all ablative variants exhibit a remarkable level of memory efficiency, as they consume less than 16GB of GPU memory. This observation is consistent with the \textbf{memory efficiency} advantage highlighted in Section~\ref{sec:Advantages}.

\paragraph{Effects of Combining M2IST with LoRA} To evaluate the extensibility of our M2IST, we combine it with the widely used PETL method, LoRA \cite{hu2021lora}. 

As shown in Table \ref{Table:mist+lora}, using LoRA alone leads to performance degradation. This is because LoRA integrates two pairs of tunable low-rank decomposed weight matrices into each encoder layer of the Vision Encoder and Language Encoder separately, without any multi-modality interaction during the feature extraction stage. Combining M2IST with LoRA performs comparably to or even better than using M2IST alone, demonstrating the extensibility of our approach. However, as discussed in Section~\ref{sec:Advantages}, such a combination may significantly increase the number of updated parameters and GPU memory usage, thus undermining the memory efficiency advantage of M2IST. Therefore, our M2IST achieves the optimal performance and efficiency trade-off in this comparison.

\subsection{Qualitative Results}
\label{sec:Qualitative}

To investigate the impact of cross-modality interaction facilitated by M3ISA, we visualize the attention scores between the \texttt{[REG]} token and other visual tokens in the V-L Encoder, where highlighted regions represent areas with higher attention scores. We compare M3ISA (referred to as "With Interaction") with its variant presented in Table~\ref{Table:ablation on mixing} (referred to as "Without Interaction") across various scenarios to assess their effectiveness in understanding challenging cases, such as human action recognition and spatial relation reasoning, as shown in Fig.~\ref{fig:visualizations}. 

It is clear that M3ISA effectively addresses most of the diverse REC cases, particularly in spatial relations. For instance, in the fifth case, the "With Interaction" approach successfully directs attention to the front region associated with the referring sentence \textit{"front pizza"}. This is especially evident in the last case, where "With Interaction" focuses more on the area corresponding to the referring sentence \textit{"right person in air"}, while "Without Interaction" fails to identify the correct region. These visualization results demonstrate that the enhanced cross-modality interaction facilitated by the IEA enables effective comprehension of complex semantic information.

However, both approaches show sub-optimal performance when handling images with \textbf{\textit{occluded}} or \textbf{\textit{crowded}} objects, as seen in cases 3, 4, and the last example. Notably, in case 3, where the target region (\textit{"girl falling down"}) is occluded by another person, both variants incorrectly allocate attention to the occluding area. Similarly, in the final case (\textit{"guy in middle"}), our method partially attends to neighboring regions. These cases collectively indicate limitations in local visual feature modeling."

\section{Conclusion}
\label{sec:Conclusion}
In this paper, we introduce Multi-Modal Interactive Side-Tuning (M2IST), an efficient tuning method designed for referring expression comprehension. Based on this framework, we introduce Mixture of Multi-Modal Interactive Side Adapters (M3ISA) to efficiently transfer pre-trained single-modality knowledge and facilitate cross-modality interaction between vision and language encoders. During fine-tuning, we freeze the pre-trained vision-language foundation models and update M3ISAs on side networks, achieving efficient tuning for REC. By updating only 3.14M encoder parameters (2.11\% of full fine-tuning) and using 15.44GB of GPU memory (39.61\% of full fine-tuning), M2IST achieves competitive performance compared to full fine-tuning methods and outperforms other PETL methods across three benchmarks.

\section{Limitations and Future Works}
\label{sec:Limiations}

In this work, we implement our M2IST on the mainstream transformer-based architecture for referring expression comprehension, comprising a pre-trained Vision Encoder and Language Encoder. However, our method shows sub-optimal performance in occluded or crowded scenes for REC. Future work will explore incorporating convolutional adapters to enhance local visual feature modeling to address such limitation.

Additionally, traditional transformer-based models demonstrate notable limitations when dealing with REC tasks in complex scenarios that demand sophisticated reasoning capabilities. Given the remarkable advancement of multi-modal large language models (MLLMs) in recent years, applying M2IST to MLLMs like LISA~\cite{Lai2024:LISA} and GLaMM~\cite{hanoona2023GLaMM} could potentially bridge this gap and significantly enhance their reasoning capabilities in complex referring scenarios. Furthermore, integrating M2IST with other model acceleration techniques~\cite{Han2024:FiCoCo,Liu2025:GlobalCom2} could further improve inference efficiency while maintaining the model's performance. This combined approach would not only leverage the powerful reasoning abilities of MLLMs but also achieve enhanced computational efficiency through multi-dimensional optimization.

Furthermore, since video processing often requires more computational resources, extending our M2IST to challenging multi-modal video understanding tasks, such as referring video object segmentation~\cite{Wu2022:referformer} and video grounding~\cite{Yang2022:Tubedetr}, will be our future research direction.

\bibliographystyle{IEEEtran}
\bibliography{references}

\end{document}